\def\year{2021}\relax
\documentclass[letterpaper]{article} 
\usepackage{aaai21}  
\usepackage{times}  
\usepackage{helvet} 
\usepackage{courier}  
\usepackage[hyphens]{url}  
\usepackage{graphicx} 
\usepackage{natbib}  
\usepackage{caption} 
\usepackage{bm,amsmath,amsfonts,multirow,booktabs}
\title{Roundtrip: A Deep Generative Neural Density Estimator}
\author{
        Qiao Liu\textsuperscript{\rm 1,2},
        Jiaze Xu\textsuperscript{\rm 1,2},
        Rui Jiang\textsuperscript{\rm 1}, and 
        Wing Hung Wong\textsuperscript{\rm 2}  \\
}
\affiliations{
    \textsuperscript{\rm 1} Tsinghua University \\
    \{liu-q16,xjz16\}@mails.tsinghua.edu.cn, ruijiang@tsinghua.edu.cn\\
    \textsuperscript{\rm 2} Stanford University \\ 
    \{liuqiao,jxu3,whwong\}@stanford.edu

}

\begin{document}

\maketitle

\begin{abstract}
Density estimation is a fundamental problem in both statistics and machine learning. In this study, we proposed Roundtrip as a general-purpose neural density estimator based on deep generative models. Roundtrip retains the generative power of generative adversarial networks (GANs) but also provides estimates of density values. Unlike previous neural density estimators that put stringent conditions on the transformation from the latent space to the data space, Roundtrip enables the use of much more general mappings. In a series of experiments, Roundtrip\footnote{Source code of Roundtrip was provided at https://github.com/kimmo1019/Roundtrip} achieves state-of-the-art performance in a diverse range of density estimation tasks.
\end{abstract}

\section{Introduction}

Density estimation is a fundamental problem in statistics. Let $p(\cdot)$ be a density on a $n$-dimensional Euclidean space $\mathcal{X}$. Our task is to estimate the density $p(\cdot)$ based on a set of independently and identically distributed data points $\{\textbf{x}_i\}_{i=1}^N$ drawn from this density.

Traditional density estimators such as histograms \cite{scott1979optimal,lugosi1996consistency} and kernel density estimators (KDEs \cite{rosenblatt1956remarks,parzen1962estimation}) typically perform well only in low dimension (e.g., $n$ is small). Recently, neural network-based approaches were proposed for density estimation, and yielded promising results for high dimensional problems (e.g., when each data point is an image). There are mainly two families of such neural density estimators: \textit{autoregressive models} \cite{uria2016neural,germain2015made,papamakarios2017masked} and \textit{normalizing flows} \cite{rezende2015variational,balle2015density,dinh2016density}. Autoregression-based neural density estimators decompose the density into the product of conditional densities based on probability chain rule $p(\textbf{x})=\prod_ip(x_i|\textbf{x}_{1:i-1})$. Each conditional probability $p(x_i|\textbf{x}_{1:i-1})$ is modeled by a parametric density (e.g., Gaussian or mixture of Gaussian), of which the parameters are learned by neural networks. Density estimators based on normalizing flows represent $\textbf{x}$ as an invertible transformation of a latent variable $\textbf{z}$ with known density, where the invertible transformation is a composition of a series of simple functions whose Jacobian is easy to compute. The parameters of these component functions are then learned by neural networks.

As suggested by \cite{kingma2016improved}, both of the above two types of neural density estimators can be viewed under the following general framework. Given a differentiable and invertible mapping $g: \mathbb{R}^n\to\mathbb{R}^n$ and a base density $p(\textbf{z})$, the density of $\textbf{x}=G(\textbf{z})$ can be represented using the \textit{change of variable rule} as 
\begin{equation}
p(\textbf{x})=p(\textbf{z})|\rm{det}(\textbf{J}_z)|^{-1}
\label{eq2}
\end{equation}
where $\textbf{J}_z=\frac{\partial G(\textbf{z})}{\partial \textbf{z}^T}$ is the Jacobian matrix of function $G(\cdot)$ at point $\textbf{z}$. Density estimation at $\textbf{x}$ can be solved if the base density $p(\textbf{z})$ is known and the determinant and inverse of Jacobian matrix are feasible to calculate. To achieve this, previous neural density estimators have to carefully design model architectures to impose constraints on the Jacobian matrix. For example, \cite{papamakarios2017masked,dinh2016density,kingma2016improved} require the Jacobian to be triangular, \cite{berg2018sylvester} constructed a low rank perturbations of a diagonal matrix as the Jacobian, \cite{karami2018generative} proposed a circular convolution where the Jacobian is a circulant matrix. These strong constraints diminish the expressiveness of neural networks which may lead to poor performance. For example, autoregressive neural density estimators based on learning $p(x_i|\textbf{x}_{1:i-1})$ are naturally sensitive to the order of the features. Moreover, the \textit{change of variable rule} is not applicable when the domain dimension differs in base density and target density. However, experience from deep generative models (e.g., GAN \cite{goodfellow2014generative} and VAE \cite{kingma2013auto}) suggested that it is often desirable to use a latent space of smaller dimension than the data space.

To overcome the limitations above, we proposed a new neural density estimator called Roundtrip. Our approach is motivated by recent advances in deep generative neural networks \cite{goodfellow2014generative,zhu2017unpaired,makhzani2015adversarial}. Roundtrip differs from previous neural density estimators in two ways. 1) It allows the direct use of a deep generative network to model the transformation from the latent variable space to the data space while previous neural density estimators use neural networks only to represent the component functions that are used for building up invertible transformation. 2) It can efficiently model data densities that are concentrated near learned manifolds, which is difficult to achieve by previous approaches as they require the latent space to have equal dimension as the data space. Importantly, we also provide methods, based on either importance sampling and Laplace approximation, for the point-wise evaluation of the density estimate. We summarize our major contributions in this study as follows.

\begin{itemize}
  \item We proposed Roundtrip as a general-purpose neural density estimator based on deep generative models. Roundtrip requires less restrictive model assumptions compared to previous neural density estimators.
  \item We provided theoretical guarantees for the feasibility of density estimation with deep generative models. Specifically, we proved that the principle in previous neural density estimators can be regarded as a special case in our Roundtrip framework (See proof in \textit{Appendix B}).
  \item We demonstrated state-of-the-art performance of Roundtrip model through a series of experiments, including density estimation tasks in simulations as well as in real data applications ranging from image generation to outlier detection.
\end{itemize}

\section{Methods}
\label{methods}
\subsection{Roundtrip overview}
The key idea of Roundtrip is to approximate the target distribution as a convolution of a Gaussian with a distribution induced on a manifold by transforming a base distribution where the transformation is learned by joint training of two GAN models (Figure \ref{fig1}).
Density estimation is an offline algorithm which is typically conducted after the two GAN models are well trained in Roundtrip. Next, we will first introduce our framework on how to model densities with deep generative networks before providing details on training strategy and model architecture.

\subsection{Density modeling in Roundtrip}
\begin{figure*}[t]
\centering
\includegraphics[width=0.8\textwidth]{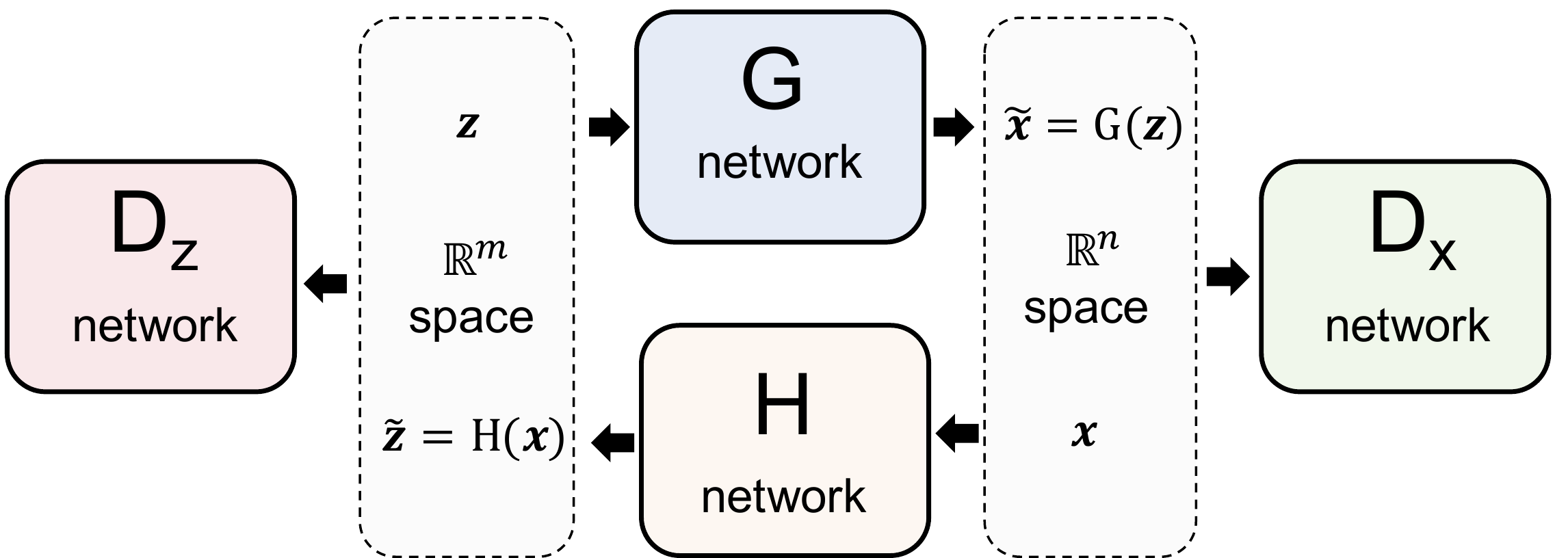} 
\caption{The overview framework of Roundtrip.}
\label{fig1}
\end{figure*}

Consider two random variables $\textbf{z}\in\mathbb{R}^m$ and $\textbf{x}\in\mathbb{R}^n$ where $\textbf{z}$ has a known density $p(\textbf{z})$ (e.g., standard Gaussian) and $\textbf{x}$ is distributed according to a target density $p(\textbf{x})$ that we intend to estimate based on $i.i.d$ observations from it. We introduced two functions $G(\cdot)$ and $H(\cdot)$ for learning an forward and backward mapping relationship between the two distributions. These two functions are learned by two neural networks (Figure~\ref{fig1}). The model architecture is similar to CycleGAN \cite{zhu2017unpaired} while we intend to exploit it for a new task of density estimation. To do this, we denote $G(\textbf{z})=\tilde{\textbf{x}}$ and $H(\textbf{x})=\tilde{\textbf{z}}$ and assume that the forward mapping error follows a Gaussian distribution
\begin{equation}
\textbf{x}=\tilde{\textbf{x}}+\bm{\epsilon}, \epsilon_i\sim N(0,\sigma^2)
\label{eq3}
\end{equation}
Typically, we set $m<n$, which means that $\tilde{\textbf{x}}$ takes values in a manifold of $\mathbb{R}^n$ with intrinsic dimension $m$. Basically, this roundtrip model utilizes $G(\cdot)$ to produce a manifold and then approximate the target density as a mixture of Gaussians where the mixing density is the induced density $p(\tilde{\textbf{x}})$ on the manifold. In what follows, we will set $p(\textbf{z})$ to be a standard Gaussian $p(\textbf{z})=(\frac{1}{\sqrt{2\pi}})^me^{-\frac{\left\|\textbf{z}\right\|_2^2}{2}}$. Based on the model assumption, $p(\textbf{x}|\textbf{z})=(\frac{1}{\sqrt{2\pi}\sigma})^ne^{-\frac{\left\|\textbf{x}-G(\textbf{z})\right\|_2^2}{2\sigma^2}}$. Then, the target density can be expressed as
\begin{equation}
p(\textbf{x})=\int p(\textbf{x}|\textbf{z})p(\textbf{z})d\textbf{z}=(\frac{1}{\sqrt{2\pi}})^{m+n}\sigma^{-n}\int e^{-\frac{v(\textbf{x},\textbf{z})}{2}}d\textbf{z}
\label{eq4}
\end{equation}
where $v(\textbf{x},\textbf{z})=\left\|\textbf{z}\right\|_2^2+\sigma^{-2}\left\|\textbf{x}-G(\textbf{z})\right\|_2^2$. The density estimation problem has been transformed to computing the integral in equation (\ref{eq4}). We will postpone model training details to section \ref{loss1}-\ref{loss3}. Assuming that $G(\cdot)$ and $H(\cdot)$ have already been well learned, we now discuss how to evaluate integral in (\ref{eq4}) by either importance sampling or Laplace approximation.

\paragraph{Importance sampling}
The simplest way to estimate (\ref{eq4}) is to use the empirical expectation by $\frac{1}{N}\sum_i^Np(\textbf{x}|\textbf{z}_i)$ where $\textbf{z}_i\sim p(\textbf{z})$. However, this is usually extremely inefficient as $p(\textbf{x}|\textbf{z})$ typically takes low values at most values of $\textbf{z}_i$ sampled from $p(\textbf{z})$. Thus we propose to sample $\textbf{z}_i$ from an importance distribution $q(\textbf{z})$ instead of the base density $p(\textbf{z})$ and use the importance-weighted estimate as
\begin{equation}
p^{IS}(\textbf{x})=\frac{1}{N}\sum_i^Np(\textbf{x}|\textbf{z}_i^q)w(\textbf{z}_i^q)
\label{eq5}
\end{equation}
where $N$ is the sample size, $w(\textbf{z})=\frac{p(\textbf{z})}{q(\textbf{z})}$ is the importance weight function, $\{\textbf{z}_i^q\}_{i=1}^N$ are $i.i.d$ samples from $q(\textbf{z})$. We propose to set $q(\textbf{z})$ to be a Student's t distribution with the center at $\tilde{\textbf{z}}=H(\textbf{x})$. This choice is motivated by the following considerations. 1) For a given $\textbf{x}$, p(\textbf{x}|\textbf{z}) is likely to be maximized at values of $\textbf{z}$ near $\tilde{\textbf{z}}=H(\textbf{x})$. 2) Student's t distribution has a heavier tail than Gaussian which provides a control of the variance of the summand in (\ref{eq5}). More details including an illustrative example of importance sampling are provided in \textit{Appendix A}. 

\paragraph{Laplace approximation}
We can also obtain an approximation to the integral in (\ref{eq4}) by Laplace's method. To achieve this goal, we expand $G(\textbf{z})$ around $\tilde{\textbf{z}}=H(\textbf{x})$ to obtain a quadratic approximation to $v(\textbf{x},\textbf{z})$, which can be represented as 
\begin{equation}
\begin{split}
\textbf{x}-G(\textbf{z})\approx\textbf{x}-G(\tilde{\textbf{z}})-\nabla G(\tilde{\textbf{z}})(\textbf{z}-\tilde{\textbf{z}})
\label{LP1}
\end{split}
\end{equation}
where $\nabla G(\tilde{\textbf{z}})\in \mathbb{R}^{n\times m}$ is the Jacobian matrix at $\tilde{\textbf{z}}$. Substitute (\ref{LP1}) into $\left\|\textbf{x}-G(\textbf{z})\right\|_2^2$, we have
\begin{equation}
\begin{split}
\left\|\textbf{x}-G(\textbf{z})\right\|_2^2 &=(\textbf{x}-G(\textbf{z}))^T(\textbf{x}-G(\textbf{z})) \\
&= \left\|\textbf{x}-G(\tilde{\textbf{z}})\right\|_2^2-2(\textbf{x}-G(\tilde{\textbf{z}}))^T\nabla G(\tilde{\textbf{z}})(\textbf{z}-\tilde{\textbf{z}}) \\ 
&+(\textbf{z}-\tilde{\textbf{z}})^T\nabla G^T(\tilde{\textbf{z}})\nabla G(\tilde{\textbf{z}})(\textbf{z}-\tilde{\textbf{z}})
\label{LP2}
\end{split}
\end{equation}
Next, We made variable substitutions as
\begin{equation}
\left\{
             \begin{array}{lr}
             \mathbf{A}=\nabla G^T(\tilde{\textbf{z}})\nabla G(\tilde{\textbf{z}}) &\in \mathbb{R}^{m\times m}\\
             \textbf{b}=\nabla G^T(\tilde{\textbf{z}})(\textbf{x}-G(\tilde{\textbf{z}})) &\in \mathbb{R}^{m}\\
             \textbf{w}=\textbf{z}-\tilde{\textbf{z}} &\in \mathbb{R}^{m}\\
             \lambda=\sigma^{-2}
             \end{array}
\right.
\label{LP3}
\end{equation}
Taking equations (\ref{LP2}) and (\ref{LP3}) into $v(\textbf{x},\textbf{z})$ in equation (\ref{eq4}) and we can get
\begin{equation}
\begin{split}
v(\textbf{x},\textbf{z}) &= \tilde{v}(\textbf{x},\textbf{w})=\left\|\textbf{w}\right\|_2^2+2\textbf{w}^T\tilde{\textbf{z}}+\left\|\tilde{\textbf{z}}\right\|_2^2 \\ &+\lambda(\left\|\textbf{x}-G(\tilde{\textbf{z}})\right\|_2^2-2\textbf{b}^T\textbf{w}+\textbf{w}^T\mathbf{A}\textbf{w}) \\
&=\textbf{w}^T(\mathbf{I}+\lambda\mathbf{A})\textbf{w}-2(\lambda\textbf{b}-\tilde{\textbf{z}})^T\textbf{w}+c_1(\textbf{x})
\label{LP4}
\end{split}
\end{equation}
where $\mathbf{I}\in \mathbb{R}^{m\times m}$ is the identity matrix and $c_1(\textbf{x})=\left\|\tilde{\textbf{z}}\right\|_2^2+\lambda\left\|\textbf{x}-G(\tilde{\textbf{z}})\right\|_2^2$.
The integral in (\ref{eq4}) $w.r.t$ $\textbf{z}$ can now be solved by constructing a multivariate Gaussian distribution $w.r.t$ $\textbf{w}$ in ({\ref{LP4}}) as the following
\begin{equation}
\begin{split}
\int e^{-\frac{v(\textbf{x},\textbf{z})}{2}}d\textbf{z} &=\int e^{-\frac{\tilde{v}(\textbf{x},\textbf{w})}{2}}d\textbf{w}=\int e^{-\frac{(\textbf{w}-\bm{\mu})^T\bm{\Sigma}^{-1}(\textbf{w}-\bm{\mu})+c_2(\textbf{x})}{2}}d\textbf{w} \\
&=e^{-\frac{c_2({\textbf{x}})}{2}}\int e^{-\frac{(\textbf{w}-\bm{\mu})^T\bm{\Sigma}^{-1}(\textbf{w}-\bm{\mu})}{2}}d\textbf{w} \\
&=e^{-\frac{c_2({\textbf{x}})}{2}}\sqrt{(2\pi)^m\rm{det}(\mathbf{\Sigma})}
\label{LP5}
\end{split}
\end{equation}
where $c(\textbf{x})=c_1(\textbf{x})-\bm{\mu}^T\bm{\Sigma}^{-1}\bm{\mu}$, $\rm{det}(\mathbf{\Sigma})$ denotes the determinant of the covariant matrix. The constructed mean and covariant matrix of the multivariate Gaussian should be 
\begin{equation}
\left\{
             \begin{array}{lr}
             \bm{\Sigma}=(\mathbf{I}+\lambda\mathbf{A})^{-1} \\
             \bm{\mu}=\mathbf{\Sigma}(\lambda\textbf{b}-\tilde{\textbf{z}})
             \end{array}
\right.
\label{LP6}
\end{equation}
Substitute (\ref{LP5}) into (\ref{eq4}) and we can get the final closed-form solution for density of point \textbf{x}
\begin{equation}
\begin{split}
p^{LP}(\textbf{x})=(\frac{1}{\sqrt{2\pi}})^n\sigma^{-n}\sqrt{\rm{det}(\bm{\Sigma})}e^{-\frac{c(\textbf{x})}{2}}
\label{eq12}
\end{split}
\end{equation}
where $\bm{\Sigma}=(\mathbf{I}+\sigma^{-2}\mathbf{J}^T_{\tilde{\textbf{z}}}\mathbf{J}_{\tilde{\textbf{z}}})^{-1}\in \mathbb{R}^{m\times m}$. $\mathbf{J}_{\tilde{\textbf{z}}}\in \mathbb{R}^{n\times m}$ is the Jacobian matrix of $G(\textbf{z})$ at $\tilde{\textbf{z}}$, $\rm{det}(\cdot)$ denotes the matrix determinant and $c(\textbf{x})$ is a scalar function. Interestingly, we note that \textit{change of variable rule} represented by (\ref{eq2}) where $G(\cdot)$ is a differentiable and invertible function is a special case in the closed-form solution (\ref{eq12}) if the following three conditions are satisfied. 1) $m=n$, 2) $H(\cdot)=G^{-1}(\cdot)$, 3) $\sigma \to 0$. The proof is given in \textit{Appendix B}. 

In the remaining part of Section \ref{methods}, we discussed how to learn  $G(\cdot)$ and  $H(\cdot)$ from given observation data.

\subsection{Adversarial training loss}
\label{loss1}
The Roundtrip model consists a pair of two GAN models. For the forward GAN mapping, $G$ aims at generating samples $\{\tilde{\textbf{x}}_i\}_{i=1}^N$ that are similar to observation data $\{\textbf{x}_i\}_{i=1}^N$ while the discriminator $D_x$ tries to discern observation data (positive) from generated samples (negative). The backward mapping function $H$ and the discriminator $D_z$ aims to transform the data distribution to approximate the base distribution in latent space. Discriminators can be considered as binary classifiers where a input data point will be asserted to be positive (1) or negative (0). The objective loss functions of the above four neural networks ($G$,$H$,$D_z$, and $D_x$) in the training process can be represented as the following
\begin{equation}
\begin{split}
\left\{
      \begin{array}{lr}
      \begin{aligned}
      \mathcal{L}_{GAN}(G) =& \mathbb{E}_{\textbf{z}\sim p(\textbf{z})}(D_x(G(\textbf{z}))-1)^2 \\
      \mathcal{L}_{GAN}(D_x) =& \mathbb{E}_{\textbf{x}\sim p(\textbf{x})}(D_x(\textbf{x})-1)^2 +\\ &\mathbb{E}_{\textbf{z}\sim p(\textbf{z})}D_x^2(G(\textbf{z}))\\
      \mathcal{L}_{GAN}(H) =& \mathbb{E}_{\textbf{x}\sim p(\textbf{x})}(D_z(H(\textbf{x}))-1)^2 \\
      \mathcal{L}_{GAN}(D_z) =& \mathbb{E}_{\textbf{z}\sim p(\textbf{z})}(D_y(\textbf{z})-1)^2+\\ &\mathbb{E}_{\textbf{x}\sim p(\textbf{x})}D_z^2(H(\textbf{x})) 
      \end{aligned}
      \end{array}
\label{eq13}
\right.
\end{split}
\end{equation}

where $\textbf{z}$ and $\textbf{x}$ are sampled from base density $p(\textbf{z})$ and data density $p(\textbf{x})$, respectively. In practice, sampling $\textbf{x}$ from data density $p(\textbf{x})$ can be regarded as a procedure of randomly sampling from $i.i.d$ obervations data with replacement. Minimizing the loss of a generator (e.g., $\mathcal{L}_{GAN}(G)$) and the corresponding discriminator (e.g., $\mathcal{L}_{GAN}(D_x)$) are somehow contradictory as the two networks ($G$ and $D_x$) compete with each other during the training process. Note that the least square loss functions we used in equation (\ref{eq13}) has been detailedly discussed in LSGAN \cite{mao2017least}. 

\subsection{Roundtrip loss}
\label{loss2}
During the training, we also aim to minimize the roundtrip loss which is defined as $\rho(\textbf{z},H(G(\textbf{z})))$ and $\rho(\textbf{x},G(H(\textbf{x})))$ where $\textbf{z}$ and $\textbf{x}$ are sampled from the base density $p(\textbf{z})$ and the data density $p(\textbf{x})$. The principle is to minimize the distance when a data point goes through a roundtrip transformation between two data domains. If $m$<$n$, this will ensure that $\textbf{x}\to H(\textbf{x})\to G(H(\textbf{x}))$ will stay close the projection of $\textbf{x}$ to the manifold induced by $G$, and $\textbf{z}\to G(\textbf{z})\to H(G(\textbf{z}))$ will stay close to $\textbf{z}$. In practice, we used $\textit{l}_2$ loss for both $\rho(\textbf{z},H(G(\textbf{z})))$ and $\rho(\textbf{x},G(H(\textbf{x})))$ as minimizing $\textit{l}_2$ loss implies the data is drawn from a Gaussian distribution \cite{mathieu2015deep}, which exactly matches our model assumption. We denoted the roundtrip loss as 
\begin{equation}
\begin{split}
\mathcal{L}_{RT}(G,H) =& \alpha\left\|\textbf{x}-G(H(\textbf{x}))\right\|_2^2+ \\ &\beta\left\|\textbf{z}-H(G(\textbf{z}))\right\|_2^2
\label{eq14}
\end{split}
\end{equation}
where $\alpha$ and $\beta$ are two constant coefficients. The idea of roundtrip loss which exploits transitivity for regularizing structured data can also be found in previous works \cite{zhu2017unpaired,yi2017dualgan}.

\subsection{Full training loss}
\label{loss3}
Combining the adversarial training loss and roundtrip loss together, we can get the full training loss for generator networks and discriminator networks as $\mathcal{L}(G,H)=\mathcal{L}_{GAN}(G)+\mathcal{L}_{GAN}(H)+\mathcal{L}_{RT}(G,H)$ and $\mathcal{L}(D_x,D_z)=\mathcal{L}_{GAN}(D_x)+\mathcal{L}_{GAN}(D_x)$, respectively. To achieve joint training of the two GAN models, we iteratively updated the parameters in the two generative models ($G$ and $H$) and the two discriminative models ($D_z$ and $D_x$), respectively. Thus, the overall iterative optimization problem in Roundtrip can be represented as 
\begin{equation}
\begin{split}
G^*,D_x^*,H^*,D_z^* =
\left\{
      \begin{array}{lr}
      arg \min\limits_{G,H}\mathcal{L}(G,H) \\
      arg \min\limits_{D_x,D_z}\mathcal{L}(D_x,D_z)&
      \end{array}
\label{eq15}
\right.
\end{split}
\end{equation}
After an iterative model training process, the learned networks $G^*$ and $H^*$ will then be used as $G(\cdot)$ and $H(\cdot)$ functions in the density estimation procedure. Note that traditional GAN-based models lack a robust stop criteria for model training. However, the training of Roundtrip can be easily evaluated by monitoring the average log likelihood of the validation set. We stop training Roundtrip when there is no further improvement of the average log likelihood on the validation set (see training details in Section \ref{setup}).

\subsection{Model architecture}
The model architecture for Roundtrip is highly flexible. In most cases, when it is utilized for density estimation tasks with vector-valued data, we used fully-connected networks for both generative networks and discriminative networks. Specifically, the $G$ network contains 10 fully-connected layers and each layer has 512 hidden nodes while the $H$ network contains 10 fully-connected layers and each layer has 256 hidden nodes. The $D_x$ network contains 4 fully-connected layers and each layer has 256 hidden nodes while the $D_z$ network contains 2 fully-connected layers and each layer has 128 hidden nodes. The leaky-Relu activation function is deployed as a non-linear transformation in each hidden layer. 

We also extended Roundtrip for estimating the density of tensor-valued data (e.g., images) by introducing a one-hot encoded class label $\textbf{y}$ as an additional input to both $G$ and $D_x$ networks in a conditional GAN (CGAN) manner \cite{mirza2014conditional}. $\textbf{y}$ will be combined in the hidden representations in $G$ and $D_x$ networks by concatenation. Compared to vector-valued data, tensor-valued data such as images will be flattened to and reshaped from vector-valued data when taken as input and output to all networks in Roundtrip, respectively. Similar to the model architecture in DCGAN \cite{radford2015unsupervised}, we used transposed convolutional layers for upsampling images from latent space for $G$ network. Besides, we used traditional convolutional neural networks for $H$, $D_x$ while $D_z$ still adopts a fully-connected network architecture. Note that Batch normalization \cite{ioffe2015batch} is applied after each convolutional layer or transposed convolutional layer (detailed hyperparameters were provided in \textit{Appendix E}).

\begin{figure*}[t]
\centering
\includegraphics[width=0.8\textwidth]{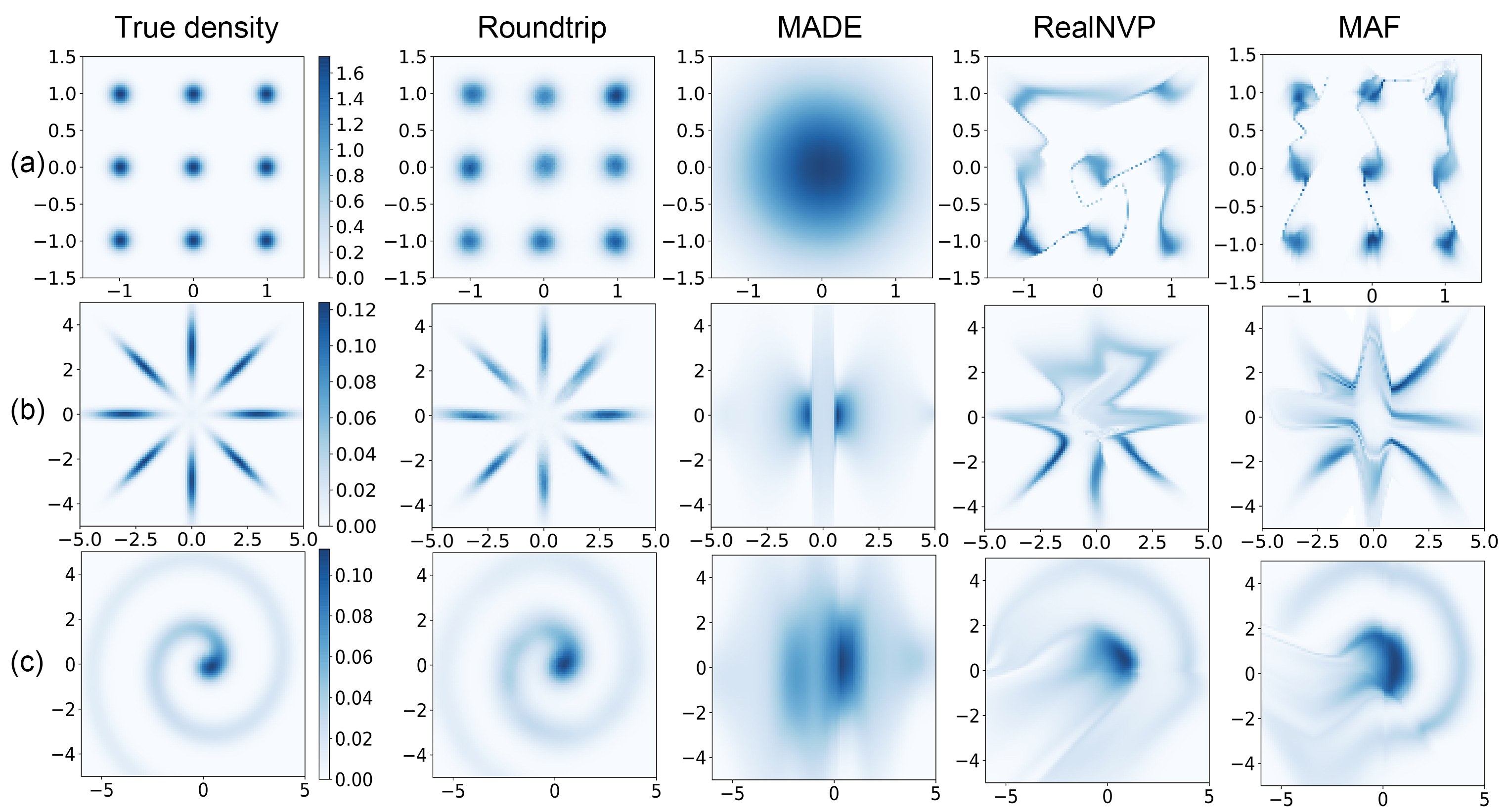} 
\caption{True density and estimated density by different neural density estimators with three simulation datasets. Density plots were shown on a 100$\times$100 grid 2D bounded region.}
\label{fig2}
\end{figure*}

\section{Results}
\subsection{Experiement setup}
\label{setup}
We test the performance of Roundtrip model in a series of experiments, including simulation studies and real data studies. In these experiments, we compared Roundtrip to the widely used Gaussian kernel density estimator as well as several neural density estimators, including MADE \cite{germain2015made}, Real NVP \cite{dinh2016density} and MAF \cite{papamakarios2017masked}. In the outlier detection experiment, we additionally compared to two commonly used outlier detection methods: One-class SVM \cite{scholkopf2001estimating} and Isolation Forest \cite{liu2008isolation}. Note that the default setting of Roundtrip model was based on the importance sampling strategy. Results of Roundtrip density estimator based on Laplace approximation are reported in \textit{Appendix C}.

The neural networks in Roundtrip model were implemented with TensorFlow \cite{abadi2016tensorflow}. In all experiments, we set $\alpha$=10 and $\beta$=10 in equation (\ref{eq14}). For the parameter $\sigma$ in our model assumption, we first pretrained the Roundtrip model for 20 epoches and selected from $\{0.01,0.05,0.1,0.2,0.4,0.5\}$ of which the value maximizes the average likelihood on validation test. Sample size $N$ in importance sampling is set to 40,000. An Adam optimizer \cite{kingma2014adam} with a learning rate of 0.0002 was used for backpropagation and updating model parameters. We stopping model training when there is no improvement on the average log-likelihood in the validation set in 10 consecutive epochs (early stopping). 

We took Gaussian kernel density estimator (KDE) as a baseline where the bandwidth is selected by Silverman's "rule of thumb" \cite{silverman1986density} or Scott's rule \cite{scott1992multivariate}. We choose the one with better results to present. The three alternative neural density estimators (MADE, Real NVP, and MAF) were implemented from \url{https://github.com/gpapamak/maf}. In outlier detection tasks, we implemented One-class SVM and Isolation Forest using scikit-learn library \cite{scikit-learn}, where the default parameters were used. To ensure fair model comparison, both simulation and real data were randomly split into a 90\% training set and a 10\% test set. For neural density estimators including Roundtrip, 10\% of the training set was kept as a validation set. The image datasets with training and test set were directly provided which require no further data split.

\begin{table*}[t]
\renewcommand\arraystretch{1.3} 
\centering
\caption{Performance of different methods on five UCI datasets. The average log likelihood (.nat) and 2 standard deviations are shown. The model with best performance is shown in bold.}\smallskip
\begin{tabular}{cccccc}
\toprule
 & AReM & CASP & HEPMASS & BANK & YPMSD \\
\midrule
KDE& 6.26$\pm$0.07&20.47$\pm$0.10&-25.46$\pm$0.03&15.84$\pm$0.12&247.03$\pm$0.61\\
MADE& 6.00$\pm$0.11&21.82$\pm$0.23&-15.15$\pm$0.02&14.97$\pm$0.53&273.20$\pm$0.35\\
Real NVP& 9.52$\pm$0.18&26.81$\pm$0.15&-18.71$\pm$0.02&26.33$\pm$0.22&287.74$\pm$0.34\\
MAF& 9.49$\pm$0.17&27.61$\pm$0.13&-17.39$\pm$0.02&20.09$\pm$0.20&290.76$\pm$0.33\\
Roundtrip& \textbf{11.74$\pm$0.04}&\textbf{28.38$\pm$0.08}&\textbf{-4.18$\pm$0.02}&\textbf{35.16$\pm$0.14}&\textbf{297.98$\pm$0.52}\\
\bottomrule
\end{tabular}
\label{table1}
\end{table*}

\subsection{Evaluation}
For simulation datasets with two dimensions, we directly visualized both true density and estimated density on a 2D bounded region. For simulation datasets with higher dimensions where the true density can be calculated, we evaluate different density estimators by calculating the Spearman (rank) correlation between true density and estimated density based on the test set. For real data where the ground truth density is not available, the average estimated density (natural log-likelihood) on the test set will be considered as a measurement. 

In the application of outlier detection, we measure performance by calculating the precision at $k$, which is defined as the proportion of correct results in the top $k$ ranks. We set $k$ to the number of outliers in the test set.

\begin{figure*}[t]
\centering
\includegraphics[width=0.8\textwidth]{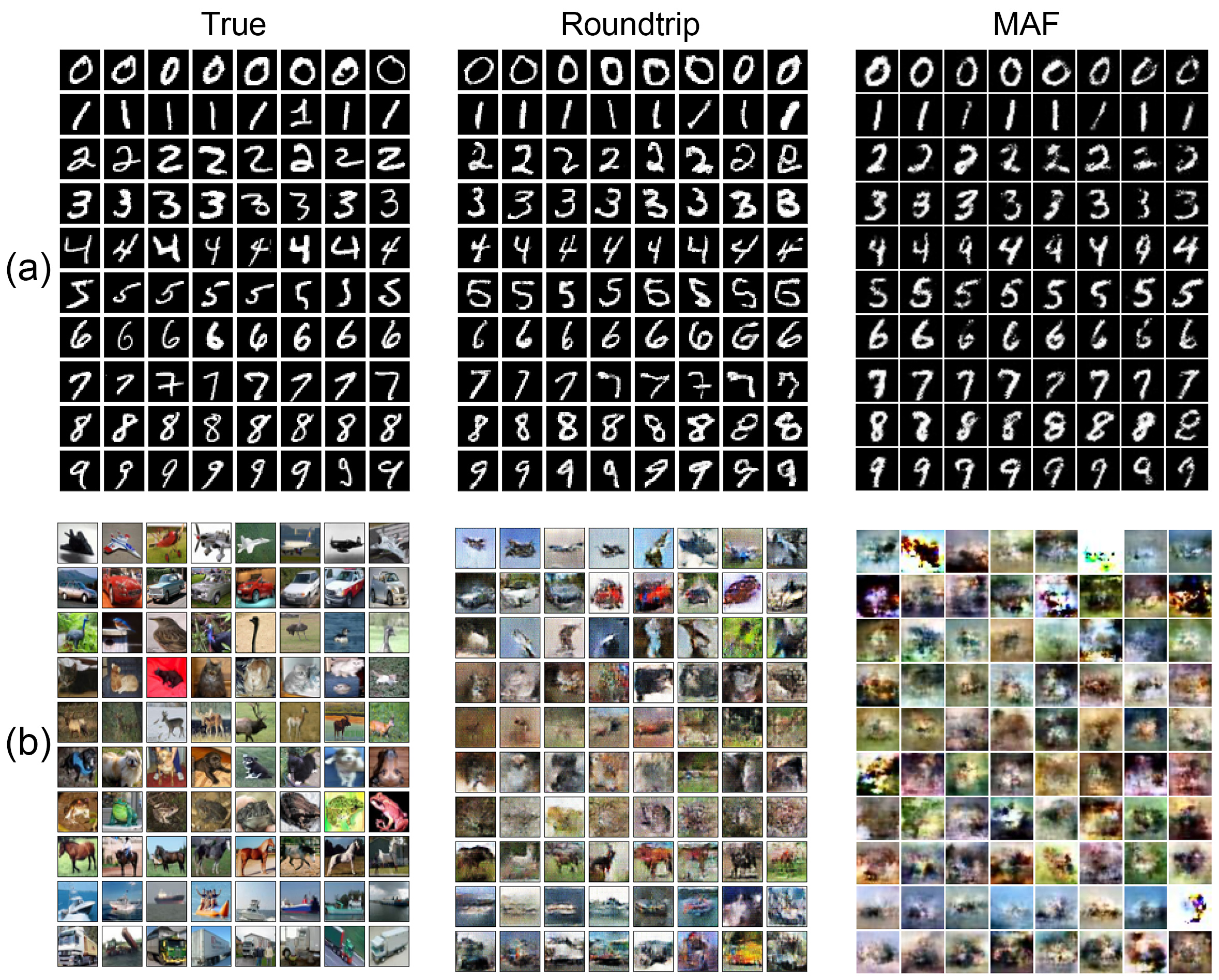} 
\caption{(a) True and generated images of MNIST. (b) True and generated images of CIFAR-10. Note that images generated by Roundtrip and MAF were sorted by decreased likelihood for each class. The results of MAF were directly collected from its original paper.}
\label{fig3}
\end{figure*}

\subsection{Simulation studies}
\label{simulation}
We first designed three 2D simulation datasets to test the performance of different neural density estimators where the truth density can be calculated.

(a) \textit{Independent Gaussian mixture}. \\$x_i \sim \frac{1}{3}(N(-1,0.5^2)+N(0,0.5^2)+N(1,0.5^2))$, $i$=1,2. 

(b) \textit{8-octagon Gaussian mixture}.\\ $\textbf{x}\sim \frac{1}{8}\sum_{i=1}^{8}N(\bm{\mu}_i,\mathbf{\Sigma}_i)$ where $\bm{\mu}_i=(3\cos\frac{\pi i}{4},3\sin\frac{\pi i}{4})$ and $\mathbf{\Sigma}_i=\bigl( \begin{smallmatrix} \cos^2\frac{\pi i}{4}+0.16^2\sin^2\frac{\pi i}{4} & (1-0.16^2)\sin\frac{\pi i}{4}\cos\frac{\pi i}{4} \\ (1-0.16^2)\sin\frac{\pi i}{4}\cos\frac{\pi i}{4} & \sin^2\frac{\pi i}{4}+0.16^2\cos^2\frac{\pi i}{4} \end{smallmatrix} \bigr)$, $i$=1,...,8.

(c) \textit{Involute}.\\ $x_1\sim N(r\sin(2r),0.4^2)$, $x_2\sim N(r\cos(2r),0.4^2)$ where $r\sim U(0,2\pi)$.

20000 $i.i.d$ points were sampled from each of the above true data distribution. After model training, we directly estimated the density in a 2D bounded region (100$\times$100 grid) with different methods (Figure \ref{fig2}). For the independent Gaussian mixture in case (a), Roundtrip clearly separates the independent components in the Gaussian mixture while other neural density estimators either failed (MADE) or contain obvious trajectory between different components (Real NVP and MAF). Roundtrip can capture a better density distribution even for the highly non-linear structure in case (c). Then we took the case (a) for a further study by increasing the dimension up to 10 (containing $3^{10}$ modes). The performance of kernel density estimator (KDE) will decrease dramatically when dimension increases. Roundtrip still achieves a Spearman correlation of 0.829 at dimension 10, compared to 0.669 of Real NVP, 0.595 of MAF, and 0.14 of KDE (See \textit{Appendix C}).

\subsection{Real data studies}
\paragraph{UCI datasets}
We collected five datasets (AReM, CASP, HEPMASS, BANK and YPMSD) from the UCI machine learning repository \cite{Dua:2019} with dimensions ranging from 6 to 90 and sample size from 42,240 to 515,345 (see more details about data description and data preprocessing in \textit{Appendix D}). Unlike simulation data, these real datasets have no ground truth for the density. Hence, we evaluated different methods by calculating the average log-likelihood on the test set. Table \ref{table1} illustrates the performance of Roundtrip and other neural density estimators. A Gaussian kernel density estimator (KDE) fitted to the training data is reported as a baseline. Roundtrip outperforms other neural density estimators by achieving the highest average log-likelihood on the test set of each dataset, which again demonstrates the superiority of our model.

\paragraph{Image datasets}

\begin{table*}[t]
\renewcommand\arraystretch{1.3} 
\centering
\caption{The precision at $k$ of different methods in three ODDS datasets.}\smallskip
\begin{tabular}{cccccc}
\toprule
 & OC-SVM & I-Forest & Real NVP & MAF & Roundtrip \\
\midrule
Shuttle& 0.953&0.956&0.784&0.929&\textbf{0.973}\\
Mammography& 0.037&\textbf{0.482}&0.474&0.407&\textbf{0.482}\\
ForestCover& 0.127&0.058&0.054&0.046&\textbf{0.177}\\
\bottomrule
\end{tabular}
\label{table2}
\end{table*}

We further applied Roundtrip model to generate images and assess the quality of the generated images by estimated density. Deep generative models have demonstrated their power in generating synthetic images. However, a deep generative model alone cannot provide quality scores for generated images. Here, we propose to use our Roundtrip method to generate images and quality score (e.g., the density of the image). We test this approach on two commonly used image datasets MNIST \cite{lecun2010mnist} and CIFAR-10 \cite{krizhevsky2009learning} where in each of the these datasets, the image comes from 10 distinct classes. Roundtrip model was modified by introducing an additional one-hot encoded class label $\textbf{y}$ to both $G$ and $D_y$ network and convolutional layers were used in $G$, $H$ and $D_x$ (see Methods). We then model the conditional density estimation by $p(\textbf{x}|\textbf{y})=\int p(\textbf{x}|\textbf{y},\textbf{z})p(\textbf{z})d\textbf{z}$ where $\textbf{y}\sim$Cat(10) denoting a categorical distribution with 10 distinct classes. We use this modified Roundtrip model to simultaneously generate images conditional on a class label and compute the within class density of the image. The comparing neural density estimators typically require a lot of tricks, including rescaling pixel values to $[0,1]$, transforming the bounded pixel values into an unbounded logit space and adding uniform noise, to achieve images generation and density estimation. Roundtrip did not require additional transformation except for rescaling. In Figure \ref{fig3}, the generated images of each class were sorted by decreased likelihood. It is seen that images generated by Roundtrip are more realistic than those generated by MAF (which is the best among alternative neural density estimators, see Figure \ref{fig2} and Table \ref{table1}). Furthermore, the density provided by Roundtrip seems to correlate well with the quality of the generated images.

\subsection{Outlier detection}
Finally, we applied Roundtrip model to an outlier detection task, where a data point with extremely low density value is regarded as likely to be an outlier. We tested this method on three outlier detection datasets (Shuttle, Mammography, and ForestCover) from ODDS database (\url{http://odds.cs.stonybrook.edu/}). 
Each dataset is split into training, validation and test set (details of data description can be found in \textit{Appendix D}). Besides the neural density estimators, we also introduced two baselines One-class SVM \cite{scholkopf2001estimating} and Isolation Forest \cite{liu2008isolation}.
The results shown in Table \ref{table2} were based on the average precision of three independent running of each algorithm. Roundtrip achieves the best or comparable results in different outlier detection tasks. Especially in the last dataset ForestCover, in which the outlier percentage is only 0.9\%, Roundtrip still achieves a precision of 17.7\% while the precision of other neural density estimators is less than 6\%.

\section{Discussion}
We proposed Roundtrip as a new neural density estimator based on deep generative models. Unlike prior studies which focus on modeling the invertible transformation between a base density and the target density, where the parameters of the components functions are learned by neural networks, Roundtrip allows the direct use of a deep generative network to model the transformation from the latent variable space to the data space. In the meanwhile, the \textit{change of variable rule} used by previous methods requires equal dimension in the base density and the target density. Roundtrip provides a more flexible transformation between the base density and target density. To the best of our knowledge, Roundtrip is the first work to tackle the general-purpose density estimation problem with deep generative neural networks (e.g., GAN). In a series of experiments, Roundtrip outperforms previous neural density estimators in a variety of density estimation tasks, including simulation/real data studies and an outlier detection application. We also demonstrated the high flexibility in Roundtrip as it can be either used for estimating density in vector-valued data and tensor-valued data (e.g., images).

Density estimation aims at obtaining accurate densities of given \textit{i.i.d} data points. Deep generative models, such as GANs, typically focus on data generative where the density is somehow implicitly learned. Our work provides a new perspective of borrowing the generation ability of deep generative models for accurately evaluating densities. 

\section{Ethical impact statement}
This work does not present any foreseeable societal consequence.

\bigskip

\bibliographystyle{aaai21} \bibliography{ref.bib}

\clearpage

\section*{Appendix A}
We used importance sampling to get numeric result of $\int p(\textbf{x}|\textbf{z})p(\textbf{z})d\textbf{z}$. One key problem is to choose an appropriate importance distribution $q(\textbf{z})$. In Roundtrip model, we chose $q(\textbf{z})$ as student's t distribution with center at $H(\textbf{x})$. $p(\textbf{x}|\textbf{z})$ always takes optimal maximum value at $\tilde{\textbf{z}}=H(\textbf{x})$ as 
\begin{equation}
\begin{split}
p(\textbf{x}|\tilde{\textbf{z}})=(\frac{1}{\sqrt{2\pi}\sigma})^ne^{-\frac{\left\|\textbf{x}-G(\tilde{\textbf{z}})\right\|_2^2}{2\sigma^2}}=(\frac{1}{\sqrt{2\pi}\sigma})^ne^{-\frac{\left\|\textbf{x}-G(H(\textbf{x}))\right\|_2^2}{2\sigma^2}}
\label{eqs1}
\end{split}
\end{equation}
We can see that minimizing roundtrip loss $\rho(\textbf{x},G(H(\textbf{x})))$ in section 2.4 is equivalent to maximizing $p(\textbf{x}|\tilde{\textbf{z}})$. This is also the reason we want to impose a roundtrip loss during the training process.

To make the importance sampling strategy more understandable, we illustrated an example based on the simulation study here. We take the \textit{Involute} simulation case in Section 3.3 for an example, we visualize $p(\textbf{z})$, $p(\textbf{x}|\textbf{z})$ and $q(\textbf{z})$ at the first dimension focusing on the density at the point $\textbf{x}$=(3,3) (Figure S1).

\begin{figure}[ht]
\centering
\includegraphics[width=0.40\textwidth]{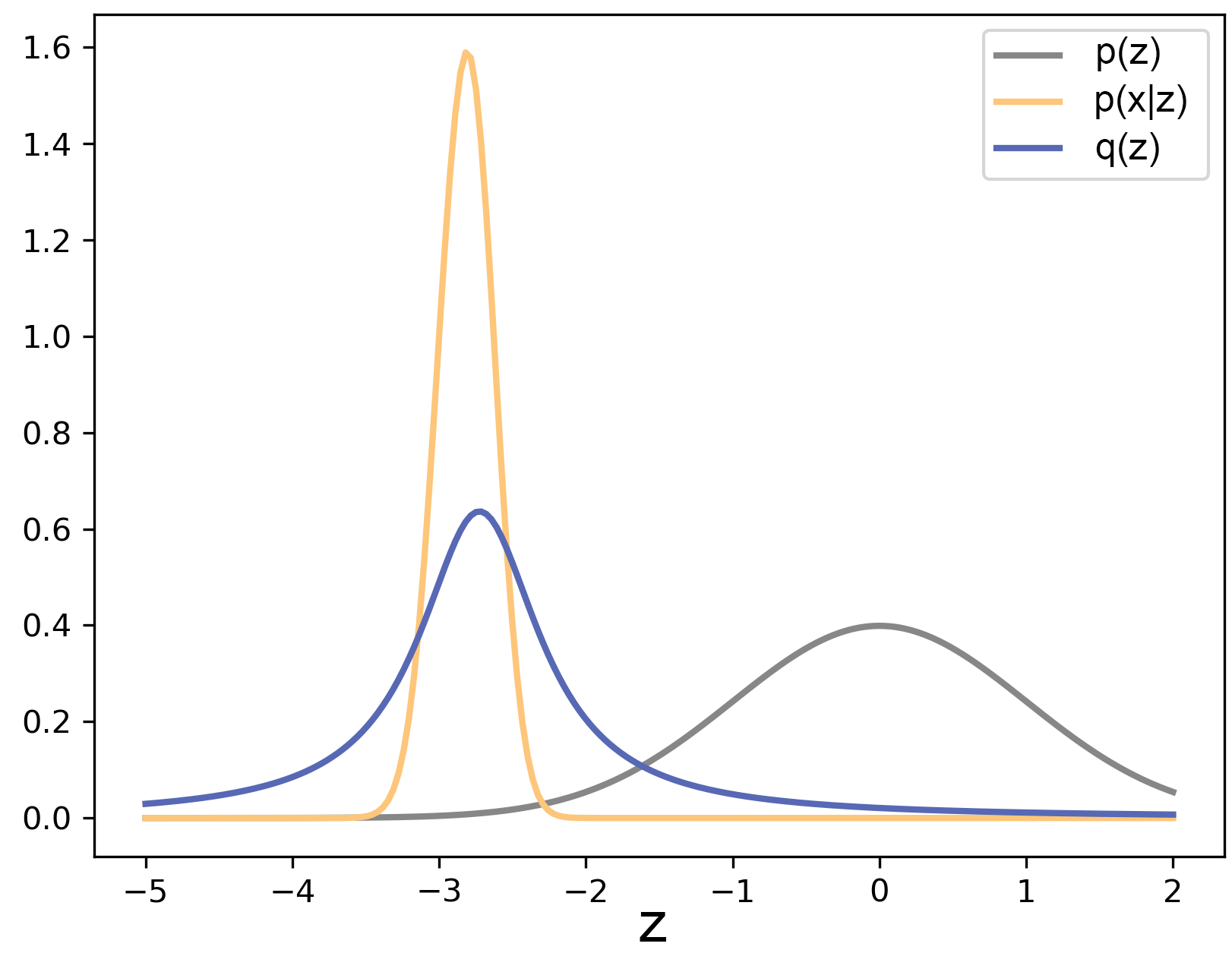} 
\begin{center}
Figure S1. Distribution of $p(\textbf{z})$, $p(\textbf{x}|\textbf{z})$ and $q(\textbf{z})$ for estimating density at point $\textbf{x}$=(3,3).
\end{center}
\label{figs1}
\end{figure}
As $p(\textbf{x}|\textbf{z})$ typically decays much faster than $p(\textbf{z})$, we chose $q(\textbf{z})$ in which the center is close to the center of $p(\textbf{x}|\textbf{z})$ as much as possible. To sum up, in the importance sampling strategy, $G(\textbf{z})$ network was used for generating samples while $H(\textbf{x})$ network was used for determining the center of importance distribution $q(\textbf{z})$.

\section*{Appendix B}
\paragraph{The \textit{Change of variable rule} as a special case}

We first rephrase the density of \textbf{x} in equation (11) in the main text as the following
\begin{equation}
\begin{split}
p(\textbf{x})&=(\frac{1}{\sqrt{2\pi}})^n\sigma^{-n}\sqrt{\rm{det}(\rm{inv}(\mathbf{I}+\sigma^{-2}\mathbf{A}))}e^{-\frac{c_2(\textbf{x})}{2}} \\
&=(\frac{1}{\sqrt{2\pi}})^n\sigma^{-n}\sqrt{\rm{det}(\sigma^2\rm{inv}(\mathbf{A}+\sigma^{2}\mathbf{I}))}e^{-\frac{c_2(\textbf{x})}{2}} \\
&=(\frac{1}{\sqrt{2\pi}})^n\sigma^{-n}\sqrt{\sigma^{2m}\rm{det}(\rm{inv}(\mathbf{A}+\sigma^{2}\mathbf{I}))}e^{-\frac{c_2(\textbf{x})}{2}}\\
&=(\frac{1}{\sqrt{2\pi}})^n\sigma^{m-n}\sqrt{\rm{det}(\rm{inv}(\mathbf{A}+\sigma^{2}\mathbf{I}))}e^{-\frac{c_2(\textbf{x})}{2}}
\label{eqs2}
\end{split}
\end{equation}
When $m=n$ and $H(\cdot)=G^{-1}(\cdot)$,then we have  $x-G(\tilde{\textbf{z}})=\textbf{x}-G(H(\textbf{x}))=\textbf{x}-G(G^{-1}(\textbf{x}))=\textbf{0}$, $\textbf{b}=\nabla G^T(\tilde{\textbf{z}})(\textbf{x}-G(\tilde{\textbf{z}}))=\textbf{0}$, $\bm{\mu}=\bm{\Sigma}(\lambda\textbf{b}-\tilde{\textbf{z}})=-\bm{\Sigma}\tilde{\textbf{z}}$ and $c_2(\textbf{x})=\left\|\tilde{\textbf{z}}\right\|_2^2-\sigma^2\tilde{\textbf{z}}^T(\mathbf{A}+\sigma^2\mathbf{I})^{-1}\tilde{\textbf{z}}$

Finally, we take the limit of $\sigma \to 0$, we have $\lim_{\sigma \to 0}c_2(\textbf{x})=\left\|\tilde{\textbf{z}}\right\|_2^2$ and
\begin{equation}
\begin{split}
\lim_{\sigma \to 0}&\sqrt{\rm{det}(\rm{inv}(\mathbf{A}+\sigma^{2}\mathbf{I}))}=\sqrt{\rm{det}(\rm{inv}(\mathbf{A}))}=\sqrt{\rm{det}(\rm{inv}(\mathbf{J}_{\tilde{\textbf{z}}}^T\mathbf{J}_{\tilde{\textbf{z}}}))} \\
&=\sqrt{\rm{det}(\mathbf{J}^{-T}_{\tilde{\textbf{z}}}\mathbf{J}^{-1}_{\tilde{\textbf{z}}})}=|\rm{det}(\mathbf{J}^{-1}_{\tilde{\textbf{z}}})|=|\rm{det}(\frac{\partial G(\tilde{\textbf{z}})}{\partial \tilde{\textbf{z}}^T})|^{-1} \\
\label{eqs3}
\end{split}
\end{equation}
So when $m$=$n$ and $H(\cdot)=G^{-1}(\cdot)$, then $\lim_{\sigma \to 0}p(\textbf{x})=(\frac{1}{\sqrt{2\pi}})^ne^{-\frac{\left\|\tilde{\textbf{z}}\right\|_2^2}{2}}|\rm{det}(\frac{\partial G(\tilde{\textbf{z}})}{\partial \tilde{\textbf{z}}^T})|^{-1}=p(\tilde{\textbf{z}})|\rm{det}(\frac{\partial G(\tilde{\textbf{z}})}{\partial \tilde{\textbf{z}}^T})|^{-1}$.

So we proved that under the three conditions (1) $m=n$, (2) $H(\cdot)=G^{-1}(\cdot)$, (3) $\sigma \to 0$, the proposed Laplace approximation is degraded into the \textit{Change of variable rule} which is the principle of previous neural density estimators. Our Laplace approximation approach can be considered as an extension of the \textit{Change of variable rule} which requires equal dimension in base density and target density.

\begin{figure}[ht]
\centering
\includegraphics[width=0.45\textwidth]{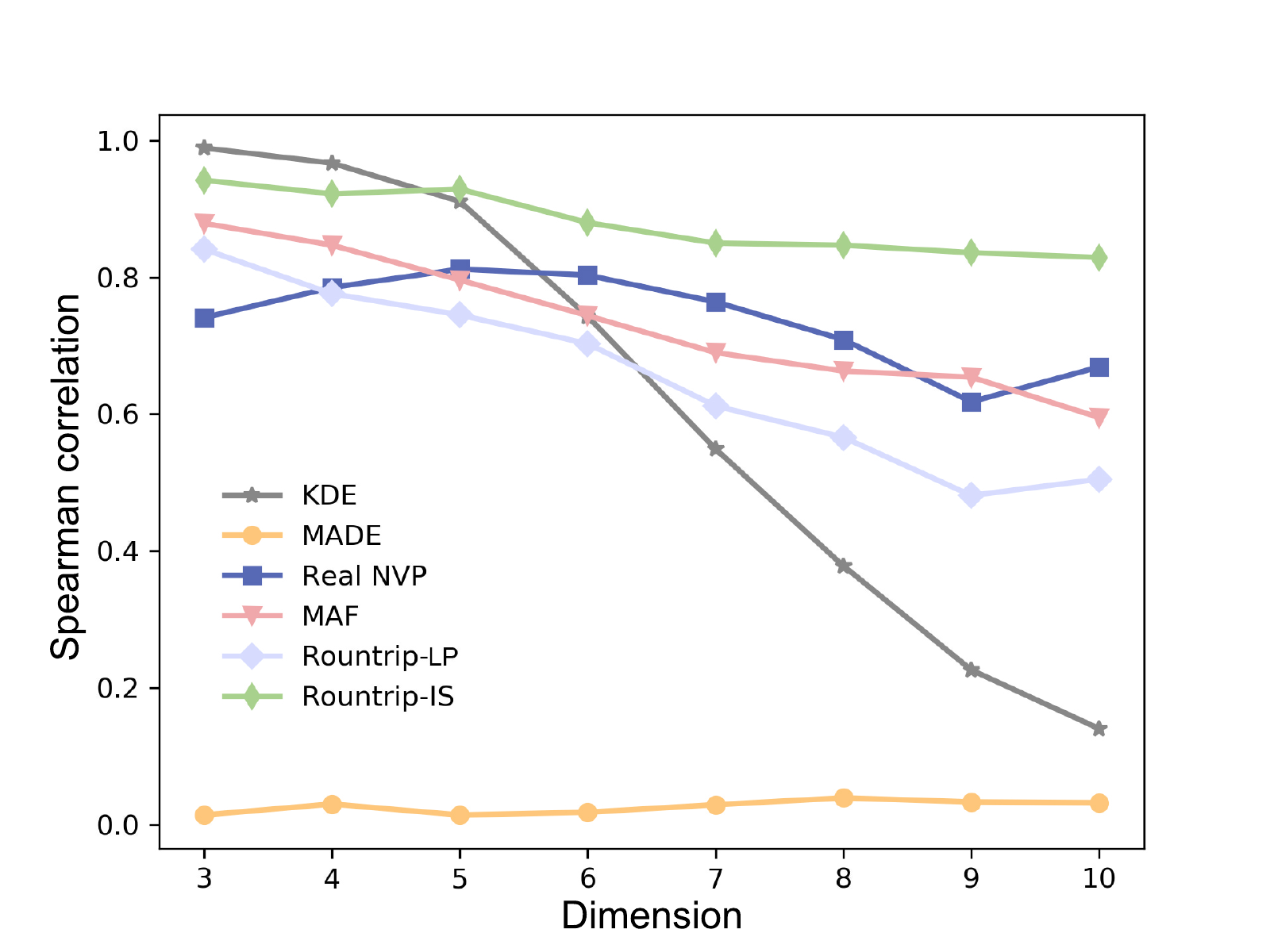} 
\begin{center}
Figure S2. Performance of different density estimators at different dimensions.
\end{center}
\label{figs2}
\end{figure}
\section*{Appendix C}

\begin{table*}[ht]{}
\centering
\begin{center}
\begin{center}
Table S1. Dimension and sample size of UCI/Image datasets
\end{center}
\end{center}
\begin{tabular}{ccccccc}
\toprule
\multirow{2}{*}{Dataset} &
\multirow{2}{*}{Domain} &
\multirow{2}{*}{Dim(\textbf{z})} &
\multirow{2}{*}{Dim(\textbf{x})} &
\multicolumn{3}{c}{Sample size} \\
\cline{5-7}
  & & & &Train & Validation & Test \\
\midrule
AReM & Social science &3 & 6 & 34215 & 3801 & 4223\\
CASP & Chemistry & 5& 9 & 37042 & 4115 & 4573\\
HEPMASS & Physics & 8& 21 & 315123 & 35013 & 174987\\
BANK & Finance & 8& 17 & 36621 & 4069 & 4521\\
YPMSD & Audio & 20 &  90 & 417430 & 46381 & 51534\\
MNIST & Image & 100&784 & 50000 & 10000 & 10000\\
CIFAR-10 & Image &100 & 3072 & 45000 & 5000 & 10000\\

\bottomrule
\end{tabular}
\end{table*}

we took the case (a) independent Gaussian mixture for a further study by increasing the dimension up to 10 (containing $3^{10}$ modes). The Spearman correlation between estimated density and true density of the test set is calculated and shown in Figure S2. The kernel density estimator (KDE) performs comparable or even better when the dimension is less than 5. But the performance of KDE decreases sharply when the dimension is larger than 5. Our Roundtrip model with the importance sampling (Roundtrip-IS) strategy can achieve a consistently better performance than other neural density estimators at different dimensions. We also note that the performance of Roundtrip model with Laplace approximation (Roundtrip-LP) outperforms MADE but not as good as MAF and Real NVP in most cases (Figure S2). 

Although we provided theoretical guarantees on the approximation solution, the success of Roundtrip-LP requires that the high order terms in equation (5) in the main text is negligible, which may introduce additional bias when estimating density. So we reported all results of density estimation using the more robust Roundtrip-IS model (default setting) as the result of importance sampling is unbiased.

\section*{Appendix D}
\paragraph{UCI and image datasets}
We provided detailed descriptions about the data description and preprocessing of all datasets that were used in our study.

\textit{AreM}. The Activity Recognition system based on Multisensor data fusion (AReM) \citep{palumbo2016human} dataset contains temporal data from a Wireless Sensor Network worn by an actor performing the activities: bending, cycling, lying down, sitting, standing, walking. The time-domain features including 3 mean values and 3 standard deviations were collected from the multisensor system during a period of time. Although it is time-series data but we treat it as if each example was drawn from an $iid$ distribution from the target distribution. Then raw data was first applied a feature scaling through a min-max normalization and then randomly split into 90\% training set and 10\% test. Note that for neural density estimators, 10\% of the training set will be kept for validation.

\textit{CASP}. The CASP dataset contains the physicochemical properties of the protein tertiary structure. Each example denotes an individual residue which has 9 features, including total surface area, non-polar exposed area, fractional area of exposed non-polar residue, fractional area of exposed non-polar part of the residue, molecular mass weighted exposed area, Euclidian distance, secondary structure penalty and spacial distribution constraints (N.K Value). The same data normalization and split were used as AreM dataset.

\textit{HEPMASS}. HEPMASS \citep{baldi2016parameterized} dataset describes the particle collisions signatures of exotic particles in high energy physics. We preprocessed this dataset following the same strategy as \citep{papamakarios2017masked}. Examples from the "1000" dataset were collected where the particle mass is 1000 and five features were removed due to too many reoccurring values.

\textit{BANK}. BANK dataset \citep{moro2014data} is related to a marketing campaign of a Portuguese banking institution where the goal is to predict whether the client will subscribe a deposit. The label encoding was used for discrete features in the raw data with values between 0 and \verb+n_classes+. Then a uniform noise of $(-0.2,0.2)$ was added to each feature. At last, the same data normalization and split were used as AreM dataset.

\textit{YPMSD}. YPMSD (\url{http://millionsongdataset.com/}) is a dataset that contains the audio features of songs from different years ranging from 1922 to 2011. Each song has 90 features which relate to 12 timbre average and 78 timbre covariance. The same data normalization and split were used as AreM dataset.

The descriptions of the five UCI datasets and the two image datasets (MNIST and CIFAR-10), including feature dimension and sample size, were summarized in Table S1. 

\begin{table*}[t]{}
\centering
\begin{center}
\begin{center}
Table S2. Dimension and sample size of ODDS datasets
\end{center}
\end{center}
\begin{tabular}{ccccccc}
\toprule
\multirow{2}{*}{Dataset} &
\multirow{2}{*}{Dim(\textbf{z})} &
\multirow{2}{*}{Dim(\textbf{x})} &
\multirow{2}{*}{Outliers(\%)} &
\multicolumn{3}{c}{Sample size} \\
\cline{5-7}
  & & & &Train & Validation & Test \\
\midrule
Shuttle & 3 & 9 & 7 & 39770 & 4418 & 4909\\
Mammograph & 3 & 6& 2.32 & 9059 & 1006 & 1118\\
ForestCover & 4 & 10& 0.9 & 231700 & 25744 & 28604\\

\bottomrule
\end{tabular}
\end{table*}

\paragraph{ODDS datasets} We provided detailed descriptions about the ODDS datasets used in this study.

\textit{Shuttle}. Shuttle (\url{http://odds.cs.stonybrook.edu/shuttle-dataset/}) dataset contains 9 numerical features. The smallest five classes, i.e. 2, 3, 5, 6, 7 are combined to form the outliers class, while class 1 forms the inlier class. Data for class 4 is discarded. All inlier and outlier data were first mixed together and then randomly split into 90\% training set and 10\% test set. For neural density estimators, 10\% of the training set were kept for validation. 

\textit{Mammography} Mammography (\url{http://odds.cs.stonybrook.edu/mammography-dataset/}) dataset describes the characteristics of 260 calcifications. The minority class of calcification is considered as an outlier class and the non-calcification class as inliers. The same data split strategy was used for Shuttle dataset.

\textit{ForestCover} ForestCover (\url{http://odds.cs.stonybrook.edu/forestcovercovertype-dataset/}) dataset is used in predicting forest cover type from cartographic variables. Outlier detection dataset is created using only 10 quantitative attributes. Instances from class 2 are considered as normal points and instances from class 4 are anomalies. The same data split strategy was used for Shuttle dataset.

The descriptions of the three ODDS datasets are summarized in Table S2.

\begin{table*}[!b]{}
\centering
\begin{center}
\begin{center}
Table S3. The network architecture for conditional image generation and density estimation.
\end{center}
\end{center}
\begin{tabular}{c|c}
\toprule
\textbf{generator} $G$ & \textbf{discriminator} $D_x$ \\
\hline
Inputs $\textbf{z}\in \mathbb{R}^{100}$ and $\textbf{y}\in \mathbb{R}^{10}$  & Inputs flattened image $\in \mathbb{R}^{784}$ and $\textbf{y}\in \mathbb{R}^{10}$  \\
\hline
Concat($\textbf{z}$, $\textbf{y}$) & Reshape, $28\times28\times1$, $1\times1\times10$  \\
\hline
FC, 1024 batchnorm. LRelu& $4\times4$ conv, 32 stride 2. batchnorm. LRelu   \\
\hline
Concat(FC,\textbf{y})& Concat(Conv, \textbf{yb})  \\
\hline
FC, $7\times7\times128$ batchnorm. LRelu& $4\times4$ conv, 64 stride 2. batchnorm. LRelu  \\
\hline
Reshape, $7\times7\times128$ & Flatten, 1568\\
\hline
Concat(Reshape,\textbf{yb})& Concat(Flat, \textbf{yb})\\
\hline
$4\times4$ upconv, 64 stride 2. Sigmoid& FC, 1024 batchnorm. LRelu\\
\hline
Flatten,784 & Concat(FC, \textbf{y})\\
\hline
 & FC, 1 \\
\midrule
\textbf{generator} $H$ & \textbf{discriminator} $D_z$ \\ 
\hline
 Input flattened image $\in \mathbb{R}^{784}$ & Input $\textbf{z}\in \mathbb{R}^{100}$ \\
\hline
 Reshape,$28\times28\times1$ & FC, 128. LRelu\\
\hline
 $4\times4$ conv, 64 stride 2. LRelu & FC, 128. Batchnorm. Tanh\\
\hline
 $4\times4$ conv, 64 stride 2. LRelu & FC, 1\\
\hline
 FC, 1024& \\
\hline
 FC, 100& \\
\bottomrule

\end{tabular}
\end{table*}

\section*{Appendix E}
Take the MNIST database for an example, we provided the details of network architectures in Roundtrip model for conditional generation and density estimation which were shown in Table S3. The one-hot encoded label $\textbf{y}$ will be fed to both $G$ network and $D_x$ network. Note that $\textbf{yb}$ is a reshape of $\textbf{y}$, which is convenient for channel-wise concatenation. For CIFAR-10 database, the network architecture used in Roundtrip is exactly the same except that the image size will be $32\times32\times3$ and the hidden unit number in the second fully-connected layer of $G$ network will be $8\times8\times128$.

\end{document}